\pdfoutput=1

\documentclass[11pt]{article}

\usepackage[]{emnlp2021}

\usepackage{times}
\usepackage{latexsym}

\usepackage[T1]{fontenc}

\usepackage[utf8]{inputenc}

\usepackage{microtype}

\usepackage{graphicx}
\graphicspath{{./Figures/}}
\usepackage{multirow}
\usepackage{amsmath}
\usepackage{changepage}
\usepackage{ulem}
\usepackage{adjustbox}

\usepackage{subcaption}
\usepackage{float}
\usepackage{cleveref}
\usepackage{booktabs}
\usepackage{makecell}
\usepackage{algorithm,algorithmic}
\usepackage{todonotes}

\usepackage[inline]{enumitem}

\usepackage{ulem}
\usepackage{times}
\usepackage{latexsym}
\usepackage{bbm} 
\usepackage{latexsym,amsmath,amsfonts,amssymb,bm}

\newcommand{\shortn}{{\scshape CEAR}}
\newcommand{\longn}{Cross-Entity Aware Reranker}
\newcommand{\boldlongn}{\textbf{C}ross-\textbf{E}ntity \textbf{A}ware \textbf{R}eranker}
\newcommand{\boldshortn}{{\textbf{CEAR}}}

\usepackage{amsmath}

\newcommand{\bi}{\begin{itemize}}
\newcommand{\ei}{\end{itemize}}
\newcommand{\BE}{\begin{enumerate}}
\newcommand{\EE}{\end{enumerate}}

\newcommand{\ie}{\mbox{\it i.e.}}

\def\complex{ComplEx}
\def\rotatE{RotatE}

\def\bs{\bm{e_s}}
\def\br{\bm{w_r}}
\def\bo{\bm{e_o}}

\def\boi{\bm{e_{o_i}}}
\def\boj{\bm{e_{o_j}}}
\def\contextAwareEmbj{\bm{c_{o_{(j)}}}}

\newcommand{\KB}{KB}
\newcommand{\kg}{\mathcal{G}}
\newcommand{\topk}{top--$k$}

\newcommand{\nl}{\mathcal{L}}
\newcommand{\relations}{\mathcal{R}}
\newcommand{\entities}{\mathcal{E}}
\newcommand{\facts}{\mathcal{T}}
\newcommand{\stageOM}{\mathcal{M}}
\newcommand{\stageOS}{\mathcal{S^M}}
\newcommand{\lm}{LM}
\newcommand{\sep}{[\text{\small SEP}]}
\newcommand{\spc}{[\text{\small SPC}]}
\newcommand{\cls}{[\text{\small CLS}]}

\newcommand{\mlp}{M}
\newcommand{\score}{\psi}
\newcommand{\loss}{L}

\newcommand{\fact}[3]{{$\langle #1, #2, #3\rangle$}}
\newcommand{\sequence}[1]{{$[#1]$}}

\def\intitle{CEAR: Cross-Entity Aware Reranker for Knowledge Base Completion}
\title{\intitle}

\author{Keshav Kolluru\textsuperscript{1*}, Mayank Singh Chauhan\textsuperscript{1*}, Yatin Nandwani\textsuperscript{1}, Parag Singla\textsuperscript{1}, Mausam\textsuperscript{1} \\
\textsuperscript{1}Indian Institute of Technology, Delhi, India \\
\texttt{\{keshav.kolluru}, \texttt{mayanksingh2298}, \texttt{yatin.nandwani\}@gmail.com}, \\ 
\texttt{\{parags}, \texttt{mausam\}@cse.iitd.ac.in}
}

\begin{document}
\maketitle

\begin{abstract}
Pretrained language models like BERT are known to be effective in storing factual knowledge about the world. This knowledge can be used to augment Knowledge Bases, which are often incomplete. 
However, prior attempts at using BERT for the task of Knowledge Base Completion (KBC) has resulted in performance \textit{worse} than the embedding based techniques that only use the graph structure. In this work we develop a novel model, \longn~(\shortn), that uses BERT to re-rank the output of existing KBC models.
Unlike prior works that score each entity independently, \shortn~ jointly scores the \topk~entities obtained from embedding based KBC models, using \textit{cross-entity} attention in BERT.
\shortn~ achieves a new state of art for the OLPBench dataset.
\end{abstract}

\section{Introduction}
\label{sec:intro}
Knowledge Bases (KBs) contain manually/semi-automatically curated assertions of the format
\fact{subject \ entity}{relation}{object \ entity}.
They are used for incorporating factual knowledge in NLP applications like Question-Answering \cite{das&al17} and Dialogue Generation \cite{raghu&al18}. However, KBs are often incomplete and the task of Knowledge Base Completion (KBC) involves discovering new links between entities. 

Pretrained language models (\lm s) have shown to memorize factual knowledge present in the vast amounts of text that is used for pretraining \cite{petroni&al19,jiang&al20}. In contrast to KBs where factual knowledge is explicitly stored in the form of triples, pretrained \lm s store it latently in their model parameters. Recent works have shown that augmenting language models with KBs is an effective way to improve downstream performance on knowledge intensive tasks \cite{verga&al20}. However, augmenting KBs with language models has not shown similar success. Models such as KG-BERT \cite{yao&al19} which use pre-trained language models for KBC still under-perform purely embedding based models like \complex\ \cite{trouillon2016complex} and \rotatE\ \cite{sun2019rotate} which only use the graph structure.

To remedy this, we present \boldlongn~(\boldshortn), a 2-stage KBC model that uses embedding based models in Stage-1 (\complex, \rotatE) and pre-trained \lm s in Stage-2 (BERT). 
The \topk~entities from Stage-1 embedding-based model are passed as input in the form of a string to the Stage-2 \lm~for re-ranking.
The string is created by concatenating the surface form of the query with the surface form of the Stage-1 \topk~entities (taken in the order of decreasing rank). 
Contextual BERT embeddings for words in an entity are pooled to create its \textit{cross-entity aware} embedding.
Finally, these are passed through an MLP to get the score of each entity which is used to re-rank them. 

Thus, \shortn~uses \begin{enumerate*}[label=(\arabic*)]
    \item BERT pre-trained knowledge,
    \item cross-entity attention and
    \item ranked entities from embedding-based KBC models
\end{enumerate*}    
 to achieve strong link prediction performance. It establishes a new state of art in FB15K-237, with HITS@1 of 42.2, 10.1 pts higher than prior models \cite{stoica&al20}. In the task of Open Link Prediction \cite{broscheit2020can}, \shortn~achieves 7.4 HITS@1, compared to 2.1 of prior models.\footnote{The code and models will be released}  
 

\begin{figure*}[htp]
\includegraphics[height=4.7cm,width=0.9\textwidth]{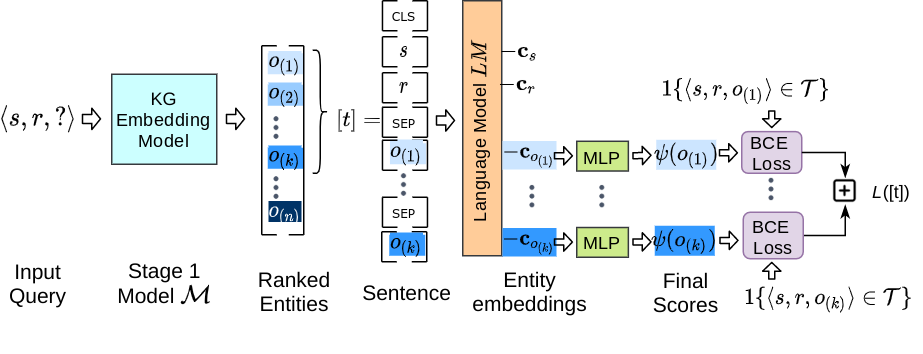}
\centering 

\hfill
\vspace*{-7ex}
\caption{The two stage architecture. Stage 1 model outputs top-\textit{k} entities that the Stage 2 model uses to generate contextual entity embeddings. The embeddings are passed through an MLP to get the final score for each entity.}
\label{fig:pipeline}
\end{figure*}

\section{Related Work}
\label{sec:related}
Many Knowledge Graph Embedding (KGE) methods have been proposed for the task of KBC \cite{bordes&al13,kazemi&al18,lacroix&al18,jain&al18}, which use various scoring functions to evaluate plausibility of triples. We experiment with \rotatE~ \cite{sun2019rotate} and \complex~ \cite{trouillon2016complex} which represent strong translation and multiplicative models.


With the rise of pre-trained language models in NLP, prior works have explored the use of BERT for Knowledge Base Completion \cite{yu&al20,kim&al20,shah&al20}. KG-BERT \cite{yao&al19} uses BERT to score all possible triples, formed by concatenation of query with each entity in the KB. Since each answer entity is scored independently, KG-BERT does not benefit from cross-entity attention. Pre-train KGE \cite{zhang&al20} uses BERT to initialize entitiy/relation embeddings used by TransE \cite{bordes&al13}.

Open Knowledge bases represent a special type of Knowledge Base which don't use a pre-defined ontology. Instead, they use fact triples generated using OpenIE systems \cite{kolluru&al20,gashteovski&al17}. The un-normalized surface forms of entities and relations makes link prediction challenging. Open Link Prediction \cite{broscheit2020can} provides a benchmark for this task, using OPIEC KB \cite{gashteovski2019opiec}, and we perform experiments on the same.



\section{Methods}
\label{sec:methods}
\noindent \textbf{Notation and Task Description}: We are given a \KB\ with relations $\relations$,  entities $\entities$ and an incomplete set of facts $\facts$.
Each fact in $\facts$ is represented as a triple \fact{s}{r}{o}, where the subject entity $s \in \entities$, is related to the object entity $o \in \entities$, via the relation $r \in \relations$. 
Given a query \fact{s}{r}{?} (or \fact{?}{r}{o}),
the task of link prediction (or KBC) is to find new facts \fact{s}{r}{o} $\notin \facts$. 
A KB can be represented as a graph, $\kg$, where for each fact \fact{s}{r}{o} $\in \facts$, the nodes representing the entities $s$ and $o$ are connected by a labeled edge $r$. 
We represent an embedding based model for KBC as $\stageOM$ and its scoring function as $\stageOS$.
$\stageOM$ learns embeddings of all the entities and relations such that $\stageOS(\bs,\br,\bo)$ is high whenever \fact{s}{r}{o} $\in \facts$. 
Here $\bs$, $\bo$ and $\br$ represent the learnt embeddings of the entities $s$, $o$ and relation $r$ respectively.

Further, we assume that we are given the surface form (or description) of the entities and relations in some natural language $\nl$, along with a pre-trained language model $\lm$, trained on some external corpus of text in the language $\nl$. 
Let \sequence{s} (or \sequence{r}) represent the sequence of words in the surface form of an entity $s \in \entities$ (or relation $r \in \relations$).
The language model $\lm$ takes a sentence, \sequence{t}, as input and generates a contextual embedding for each token in the sentence \sequence{t}.



\noindent \textbf{Motivation}: Embedding based models score all the entities independently for a given query \fact{s}{r}{?},
\ie, $\forall i \neq j$, the scores $\stageOS(\bs,\br,\boi)$ and $\stageOS(\bs,\br,\boj)$ of the entities $o_i$ and $o_j$ are computed independent of each other.
Note that such a model exploits only the structural information present in the corresponding graph.
Further, these models treat the entities and relations as atomic objects, making their score oblivious to the information present in the surface forms.

In contrast, models such as KG-BERT exploit the information in the surface forms of entities and relations using BERT. 
But, similar to embedding based models, they also score each entity independently, ignoring the relationship between the entities that may help in finding the correct entity.



\shortn\ not only exploits the benefits of both the approaches, it also overcomes their common shortcoming of scoring entities independently.
It follows a two stage approach as described below. 

\subsection{\shortn: \boldlongn}


\noindent \textbf{Stage-1: Score using an Embedding model $\stageOM$}

\noindent In the first stage, \shortn\ exploits the structural information present in the graph, $\kg$. 
It trains an embedding based model $\stageOM$, such as \complex~or \rotatE, to rank all the entities $o_j \in \entities$ based on their score $\stageOS(\bs, \br, \boj)$ for a given query \fact{s}{r}{?}.
Once such a model is trained, we pick the \topk\ answers $o_{(j)}$ for any query and pass them to Stage-2 (described below) for re-ranking them based on their surface forms and world knowledge in the pre-trained language model.
The value of $k$ depends on the capacity of the pre-trained language model $\lm$, which usually restricts the maximum number of tokens in a sentence that it can process.\footnote{BERT has a 512 word-pieces limit}

\noindent \textbf{Stage-2: Re-rank using Language Model $\lm$}
For a query \fact{s}{r}{?}\footnote{Similar formulation holds for head-entity prediction}, the surface form ($[o_{(j)}], j = 1 \ldots k$) of the \topk~ entities 
retrieved from stage 1 are used along with the surface form ($[s]$ and $[r]$) of the head entity and relation to create a sentence in the natural language $\nl$.
Specifically, \sequence{t} = 
$\cls [s] \spc [r] \sep [o_{(1)}] \sep \ldots \sep [o_{(k)}]$,
represents such a sentence, where $\cls$ is a special token to mark the beginning of a sentence; $\spc$ 
is a special token used to separate the subject from the relation;
and $\sep$ is a special token separating different answers from each other as well as from the query.
Note that the description of an entity or relation may contain multiple tokens.


The sentence \sequence{t} is fed as input to the language model $\lm$, which generates a context-aware embedding for each token in the sentence. 
The embeddings of the tokens belonging to a candidate entity $o_{(j)}$ are mean pooled to create its final embedding, $\contextAwareEmbj$. 
Such an embedding is aware of not only the query entity and relation, but also the other plausible answers. Such \textit{cross-entity aware} embeddings make use of additional context that is helpful to answer the query \fact{s}{r}{?}.

Finally, $\contextAwareEmbj$ is passed through an MLP $\mlp$ to generate its final score, $\score(o_{(j)})$, \ie, $\score(o_{(j)}) = \mlp(\contextAwareEmbj)$. 
Thus, the \topk~ entities from Stage-1 are re-ranked based on their final scores.
The $\lm$ is trained by minimizing the standard Binary Cross Entropy (BCE) loss, $\loss([t])$ for the sentence \sequence{t}, computed  using the final scores  $\score(o_{(j)}), \forall j = 1 \ldots k$, 
\vspace{-3ex}
\begin{equation}
    \loss([t]) =  -\sum\limits_{j = 1}^k (L_p^j + L_n^j) 
\end{equation}
\noindent where $L_p^j = \mathbbm{1}\{\langle s,r,o_{(j)} \rangle \in \facts\} 
    \log(\sigma(\score(o_{(j)})))$ and 
    $L_n^j =  \mathbbm{1}\{\langle s,r,o_{(j)} \rangle \notin \facts\} \log(1 - \sigma(\score(o_{(j)})))$, and 
$\sigma$ is the standard sigmoid function.

\section{Experimental Setting}
\label{sec:experimental_setting}
\textbf{Datasets:} We consider three link prediction datasets: FB15K-237 \citep{toutanova2015observed}, WN18RR \cite{dettmers2018convolutional}, and OLPBENCH \cite{broscheit2020can}. OLPBENCH proposes multiple train sets based on test set leakage removal. We use the most difficult train data set called \textit{thorough train dataset} which contains the harshest test evidence removal. 


\begin{table}[H]
\centering
\small
\begin{tabular}{@{} l r r r r r @{}}
\toprule
Dataset & Entities & Relations & Train & Valid & Test\\ \midrule
FB15K-237 & 14,541 & 237 & 272K & 17K & 20K \\ 
WN18RR    & 40,943   & 11 & 86K & 3K & 3K     \\ 
OLPBENCH  & 2.47M & 961K & 30.6M & 10K & 10K \\ \bottomrule
\end{tabular}

\caption{Statistics of the 3 datasets used.}
\label{tab:datasetstats}
\end{table}

\noindent \textbf{Evaluation:} Link prediction performance is the average of head entity and tail entity prediction. Evaluation is done under filtered settings where the model is not penalized for ranking entities appearing with query in train and val sets, higher than the gold entity. We report MRR, HITS@N metrics.
\begin{table}[H]
\centering
\small
\begin{adjustbox}{max width=0.49\textwidth}
\begin{tabular}{@{}p{1.9cm}rrrrrrr@{}}
\toprule
 \multicolumn{1}{l}{Model} & \multicolumn{3}{c}{FB15K-237} & \multicolumn{3}{c}{WN18RR} \\ \midrule
  &
  \multicolumn{1}{r}{MRR} &
  \multicolumn{1}{r}{H1} &
  \multicolumn{1}{r}{H10} &
  \multicolumn{1}{r}{MRR} &
  \multicolumn{1}{r}{H1} &
  \multicolumn{1}{r}{H10} \\ \cmidrule{2-7}
DistMult & 0.24 & 15.5 & 41.9 & 0.43 & 39.0 & 49.0 \\
ConvE & 0.33 & 23.7 & 50.1 & 0.43 & 40 & 52 \\
ComplEx-N3 & 0.37 & 27 & 56 & 0.48 & 44 & 57 \\
CoPER:ConvE & 0.43 & 32.1 & \textbf{62.9} & 0.48 & 44 & 56.1 \\ 
KG-BERT  & -        & -        & 42.0   & -       & -       & 52.4  \\
Pre-train KGE  & 0.34    & -    & 53.4   & 0.45      & -      & \textbf{58.0}      \\ 
ComplEx & 0.32    & 23    & 51.3   & 0.47   & 42.8   & 55.5  \\
RotatE & 0.34    & 23.8    & 53.1   & 0.47   & 42.3   & 57.3  \\
\shortn(ComplEx)\textsuperscript{*}     & \textbf{0.48}    & \textbf{42.2}    & 57.9   & 0.47   & 43   & 54.3 \\ 
\shortn(RotatE)\textsuperscript{*}     & 0.45    & 38.3    & 56.7   & \textbf{0.49}   & \textbf{44.3}   & 56.5  \\ \bottomrule
\end{tabular}
\end{adjustbox}
\caption{Link Prediction on FB15K-237 and WN18RR. \textsuperscript{*}() indicates the Stage-1 model used in CEAR.}
\label{tab:ckg}
\end{table}

\noindent \textbf{Baselines:} 
For FB15K-237 and WN18RR, we compare with embedding  models such as DistMult \cite{yang2014embedding}, RotatE\footnote{\label{rotate_code}  \href{https://github.com/DeepGraphLearning/KnowledgeGraphEmbedding}{github:DeepGraphLearning/KnowledgeGraphEmbedding}} \cite{sun2019rotate}, ComplEx\textsuperscript{\ref{rotate_code}} \cite{trouillon2016complex}, ConvE \cite{dettmers&al18}, ComplEx-N3 \cite{lacroix&al18}, CoPER-ConvE \cite{stoica&al20} and BERT-based models such as KG-BERT \cite{yao&al19} and Pre-train KGE \cite{zhang&al20}. For CEAR Stage-1, we use ComplEx and RotatE. 

For OLPBENCH, we compare with the state-of-the-art ComplEx-LSTM \cite{broscheit2020can}, which uses LSTM embeddings in a ComplEx model. 
Considering the large number of target entities in OLPBENCH, we also experiment with ExtremeText \cite{wydmuch2018no}, an extreme classification model, 
which builds a hierarchical softmax tree \cite{morin2005hierarchical} over FastText  \cite{joulin2016bag} embeddings. While training ExtremeText, we enriched the query by appending it with the top-5 most frequent entities seen with the relation in training data. We use ExtremeText and Complex-LSTM as Stage-1 in \shortn.

\begin{table}
\centering
\small
\begin{tabular}{@{}llrr@{}}
\toprule
Method & H1 & H10 & H50\\ \midrule
ComplEx-LSTM & 2.1 & 7.0 & 14.6\\
ExtremeText & 6.4 & 16.3 & {\bf 26.0}\\
CEAR (ComplEx-LSTM) & 3.8 & 9.1  & 14.6 \\
CEAR (ExtremeText) & \textbf{7.4} & \textbf{17.9} & \textbf{26.0}  \\ \bottomrule 
\end{tabular}
\caption{Link Prediction performance on OLPBENCH.}
\label{tab:okg}
\end{table}

\section{Experiments}
\label{sec:experiments}
\noindent \textbf{Effectiveness of \shortn}: In \Cref{tab:ckg}, we compare MRR, HITS@1 and HITS@10 of various models and find that \shortn~outperforms multiple embedding-based as well as BERT-based KBC models. It achieves a new state of art HITS@1 of 42.2 in FB15K-237, which is 19.2 pts higher than the corresponding Stage-1 model. Note that in the Stage-2 BERT of \shortn, we use \topk=40 entities from Stage-1. We truncate their surface forms to a maximum of 10 tokens, so that the constructed input, after concatenation with special tokens, has at most 512 tokens (a limitation imposed by BERT).

KG-BERT and Pre-train KGE both use BERT as part of their models and hence have access to the same amount of pre-trained knowledge as \shortn. But we find that \shortn(ComplEx) scores 15.9 pts higher compared to HITS@10 of KG-BERT. This shows the effectiveness of the \shortn~ 2-stage architecture that utilizes cross-entity attention. 

In \Cref{tab:okg}, we find that ExtremeText performs 4.3 HITS@1 higher than the previous state of art model, LSTM-ComplEx. This demonstrates the effectiveness of modeling the task as an extreme classification problem over the 2.47 million entities. We observe consistent gains by applying Stage-2 BERT on top of both LSTM-Complex (+1 HITS@1) and ExtremeText (+1 HITS@1). Thus our final model \shortn(ExtremeText) represents a 5.3 HITS@1 gain over the current state of art model, LSTM-ComplEx. We trained the Stage-2 model with only a fraction of the training data (1M out of 30M available) as we don't observe much performance gains on adding more examples.


\noindent \textbf{Ablation}: In \Cref{tab:ablation}, we compute the performance of \shortn(Best)\footnote{ComplEx, RotatE, ExtremeText are best Stage-1 models for FB15K-237, WN18RR and OLPBENCH, respectively.}
by \begin{enumerate*}[label=(\arabic*)]
    \item replacing pretrained BERT parameters with random initialization,
    \item scoring each Stage-1 entity independently (similar to KG-BERT applied only on Stage-1 entities), and
    \item randomly shuffling \topk~Stage-1 entities before passing them to Stage-2
\end{enumerate*}. 
We find that all 3 components of \shortn~are important for achieving the final model performance. Apart from pretrained knowledge,  knowing all the \topk~Stage-1 entities (in ranked order) is crucial for the model performance. 
For example, in FB15k-237, shuffling Stage-1 entities or removing Cross-Entity attention leads to a drop in HITS@1 of 17.4, 19.0 pts. 
In \Cref{tab:s1_entities}, we find an increasing trend of HITS@1 with greater number of Stage-1 entities. 
Further, we find that the Stage-2 models introduce very few errors in already correct Stage-1 model outputs (0.6\%, 0.1\% in FB15k-237, WN18RR, respectively). We report the confusion matrix, head vs. tail performance and reproducibility checklist in the Appendix.



\begin{table}[t]
\centering
\small
\begin{adjustbox}{max width=0.49\textwidth}
\begin{tabular}{@{}llrrrrr@{}}
\toprule
Model & Dataset          & k=10    & 20   & 30    & 40     \\ \midrule
CEAR(ComplEx) & FB15K-237 & 31.9 & 37.2 & 39.9 &  \textbf{42.2} \\
CEAR(RotatE) & WN18RR & 43.9 & 44.0 & 44.1 & \textbf{44.3} \\ \bottomrule
\end{tabular}
\end{adjustbox}
\caption{H@1 with increasing top-\textit{k} Stage-1 samples.}
\label{tab:s1_entities}
\end{table}

\begin{table}
\centering
\small
\begin{tabular}{@{}lrrrrrrrr@{}}
\toprule
Model           & \multicolumn{2}{c}{FB15k-237} & \multicolumn{2}{c}{WN18RR} & \multicolumn{2}{c}{OLPB} \\ \midrule
           & H1           & H10          & H1         & H10         & H1          & H10          \\ \cmidrule{2-7}
CEAR(Best)   &    \textbf{42.2}           &    \textbf{57.9}           &    \textbf{44.3}         &  56.5            &   \textbf{7.4}           & \textbf{17.9}              \\
- Pretraining &  38.5  & 56.2   &  42.7    &    55.1  & 6.0 & 15.6               \\
- CE Attention &  23.2  & 51.3  &  41.5    &    57.3  & 5.0 & 16.3              \\
- Stage1 Ranks &  24.8    &    51.7   &    40.0  & \textbf{61.3}             &  6.2  & 16.8              \\ \bottomrule
\end{tabular}
\caption{Ablation of the best CEAR model, which shows the importance of BERT pretrained knowledge, Cross-Entity Attention and Stage-1 Entity Ranks.}
\label{tab:ablation}
\end{table}

\section{Conclusion}
We present a novel KBC model, \shortn, that uses the pretrained parameters in BERT,  information of the graph structure (using \complex, \rotatE) and global view of competing enitites (using cross-entity attention) to achieve a new state of art performance for link prediction across 3 datasets. Replacing the Stage-1, Stage-2 pipeline with an end-to-end trainable model and exploring computationally feasible methods for applying cross-entity attention over larger number of entities are future directions.


\bibliography{anthology,custom}
\bibliographystyle{acl_natbib}

\clearpage
\appendix
\section*{\intitle\\ (Appendix)}


\section{Head vs. Tail performance}

In \Cref{tab:ckg}, following standard evaluation protocols, we report performance as the average of head and tail entity prediction performance. We report their individual performances in \Cref{tab:headtail} for the Stage-1 model and the corresponding CEAR model. As observed in other KBC models, we find that the absolute tail performance is significantly higher than the head performance. We also find that the gains from our Stage-2 BERT model is higher for tail compared to head prediction in FB15k-237 (23.3 vs. 16.6 HITS@1) and WN18RR (2.1 vs. 1.7 HITS@1). However the same trend does not hold for gains on OLPBENCH.

\begin{table}[H]
\centering
\small
\begin{tabular}{@{}lrrrrrrrr@{}}
\toprule
Dataset           & \multicolumn{2}{c}{Head Entity} & \multicolumn{2}{c}{Tail Entity} \\ \midrule
           & Stage-1           & CEAR          & Stage-1         & CEAR \\ \cmidrule{2-5}
FB15K-237 &  14.4  & 31.0   &  31.6    &    54.3                 \\
WN18RR &  40.8  & 42.5  &  43.8    &    45.9                \\
OLPBENCH &  4.8    &    5.9   &    8.1  & 8.6              \\ \bottomrule
\end{tabular}
\caption{H@1 of Stage-1 model and corresponding CEAR(Best) model for head and tail entity prediction}
\label{tab:headtail}
\end{table}

\section{Confusion Matrix}
In \Cref{tab:confusion} we report the number of test queries where CEAR(Best) improves the corresponding Stage-1 prediction, i.e, changes from an \textit{incorrect} entity to a \textit{correct} entity) as well as the number of test queries where CEAR changes Stage-1 prediction from correct to incorrect entity. 
For this purpose, we build a confusion matrix with respect to the change in the predicted entity before and after application of Stage-2 model.   
The complete set of test queries (both head and tail) are divided into one of four cases - 00, 01, 10 and 11. 
Case 00 indicates both Stage-1 and Stage-2 prediction were incorrect. 
Case 01 indicates stage 1 incorrect, while stage 2 correct. 
Case 10 indicates stage 1 correct and stage 2 incorrect and Case 11 indicates both stage 1 and stage 2 correct. 
It is interesting to observe that for datasets FB15K-237 and WN18RR the fraction of examples which degrade are very few, 0.6\% and 0.1\% respectively.
\begin{table}[H]
\centering
\small
\begin{tabular}{@{}lrrrr@{}}
\toprule
Dataset           & 00 & 01 & 10 & 11 \\ \midrule
FB15K-237 &  23202  & 8311   &  251    &    9168                 \\
WN18RR &  3491  & 125  &  7    &    2645                \\
OLPBENCH &  19123    &    572   &    405  & 922              \\ \bottomrule
\end{tabular}
\caption{Classifying test queries to understand the effectiveness of CEAR model }
\label{tab:confusion}
\end{table}

\begin{table*}
\centering
\small
\begin{tabular}{@{}lrrrrrrrrrrr@{}}
\toprule
Model           & \multicolumn{3}{c}{FB15k-237} & \multicolumn{3}{c}{WN18RR} &  \\ \midrule
           & Train time           & Inference time          & Parameters   & Train time           & Inference time          & Parameters  \\ \cmidrule{2-7}
Stage 1   &    4.5 hrs           &    0.5 mins           &    30M         &  3 hrs            &   0.5 mins           & 41M                \\
CEAR(Best)   &    20 hrs           &    8 mins           &    110M         &  15 hrs            &   1.5 mins           & 110M      \\
\bottomrule
\end{tabular}
\caption{Parameters and Training/Inference time on FB15k-237 and WN18RR}
\label{tab:params}
\end{table*}

\begin{table*}
\centering
\small
\begin{tabular}{@{}lrrr@{}}
\toprule
Model           &  \multicolumn{3}{c}{OLPBENCH} \\ \midrule
           & Train time           & Inference time          & Parameters\\ \cmidrule{2-4}
Stage 1  & 4 hrs            &   0.5 mins           & 14644M              \\
CEAR(Best) & 25 hrs            &   1.5 mins           & 110M     \\
\bottomrule
\end{tabular}
\caption{Parameters and Training/Inference time on OLPBENCH}
\label{tab:params2}
\end{table*}

\section{Performance across Stage-1 ranks}

\begin{figure}[htp]
\includegraphics[scale=0.4]{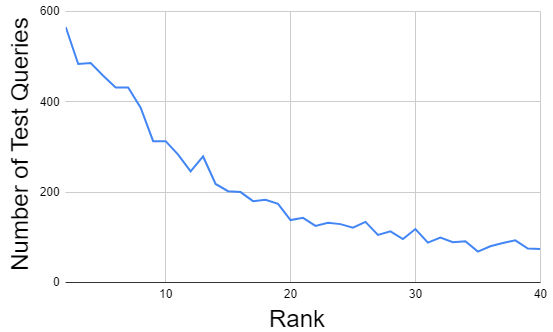}
\centering 
\hfill
\caption{Number of test queries where CEAR(Best) predicts \textit{ith} ranked Stage-1 entity for FB15k-237.}
\label{fig:fb15k237}
\end{figure}

\begin{figure}[htp]
\includegraphics[scale=0.4]{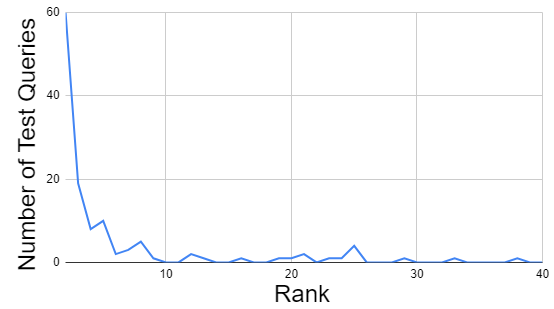}
\centering 
\hfill
\caption{Number of test queries where CEAR(Best) predicts \textit{ith} ranked Stage-1 entity for WN18RR.}
\label{fig:wn18rr}
\end{figure}

\begin{figure}[htp]
\includegraphics[scale=0.4]{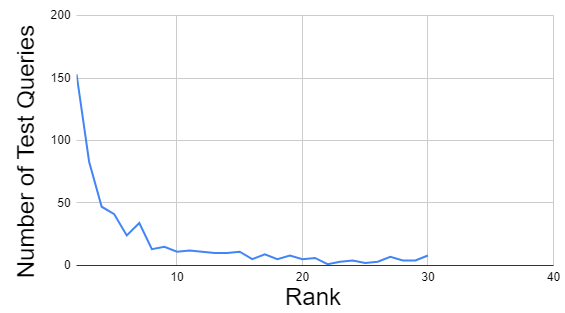}
\centering 
\hfill
\caption{Number of test queries where CEAR(Best) predicts \textit{ith} ranked Stage-1 entity for OLPBENCH.}
\label{fig:olpbench}
\end{figure}

For the Case 01 (described in the previous section) - where \shortn~corrected the Stage-1 prediction, we further divide the test queries based on the Stage-1 rank of the \shortn~predicted entity. In Figures \ref{fig:fb15k237}, \ref{fig:wn18rr}, \ref{fig:olpbench}, we plot the number of test queries where the model predicts the \textit{ith} ranked Stage-1 entity. Note that for all datasets we observe that \shortn~is biased towards predicting earlier Stage-1 samples which is not surprising given the fact that the Stage-1 models are strong KBC models which do give high ranks to correct entities. 

\section{Reproducibility Checklist}
\noindent \textbf{Compute Infrastructure}: The submitted code contains instructions to replicate the reported results. We run all of our experiments on a single GPU - NVIDIA Tesla V100 (32 GB).

\noindent \textbf{Hyper-parameters}:
We train all models for 10 epochs (5 for OLPBENCH), and save the model which gives best validation results. Learning rate is kept at 2e-5 and we use a dynamic batch size such that each batch contains a maximum of 5000 tokens.

\noindent \textbf{Validation Scores}: \Cref{tab:validation} contains the validation performance for CEAR(Best) and the corresponding Stage-1 models.

\begin{table}[H]
\centering
\small
\begin{tabular}{@{}lrrrrrrrr@{}}
\toprule
Model           & \multicolumn{2}{c}{FB15k-237} & \multicolumn{2}{c}{WN18RR} & \multicolumn{2}{c}{OLPBENCH} \\ \midrule
           & H1           & H10          & H1         & H10         & H1          & H10          \\ \cmidrule{2-7}
Stage 1   &    22.2           &    45.5           &    42.8         &  57.2            &   6.7           & 16.4              \\
CEAR(Best)   &    40.8           &    51.9           &    44.7         &  56.3            &   7.2           & 17.7              \\
\bottomrule
\end{tabular}
\caption{Validation results}
\label{tab:validation}
\end{table}

\noindent \textbf{Training/Inference Time and Number of parameters}: 
\Cref{tab:params} and  \Cref{tab:params2} contain the Training/Inference time and number of model parameters for CEAR(Best) and the corresponding Stage-1 models.

\noindent \textbf{Evaluation Metrics}:
Filtered evaluation for the task of link prediction has been done by following standard procedure as done in \href{https://github.com/dair-iitd/KBI}{KBI}. For open link prediction, appropriate modifications are done to support alternate mentions as done in \href{https://github.com/samuelbroscheit/open_knowledge_graph_embeddings}{Open Knowledge Graph Embeddings}.


\end{document}